\documentclass[letter, 10 pt, conference]{ieeeconf}

\IEEEoverridecommandlockouts  
\usepackage{cite}
\usepackage{comment}
\usepackage{amsmath, amssymb, amsfonts}
\usepackage{algorithm}
\usepackage{algpseudocode}
\usepackage{graphicx}
\usepackage{textcomp}
\usepackage{stfloats}
\usepackage{booktabs} 
\usepackage{hyperref}
\usepackage{xcolor}

\def\BibTeX{{\rm B\kern-.05em{\sc i\kern-.025em b}\kern-.08em
    T\kern-.1667em\lower.7ex\hbox{E}\kern-.125emX}}
    
\begin{document}


\title{Real-World Deployment of a 5G-Connected Edge-Controlled Aerial Robot in Industrial Subterranean Mines}


\author{Achilleas Santi Seisa$^{*}$, Emanuele Pagliari, Gerasimos Damigos, Elias Small and George Nikolakopoulos
\thanks{This work has been partially funded by the European Union, under the Grant Agreement No. 101139257 (SUNRISE-6G).}
\thanks{The authors are with the Robotics and AI Group, Department of Computer, Electrical and Space Engineering, Lule\aa\,\, University of Technology, Lule\aa\,\,}
\thanks{$^{*}$Corresponding Author's email: {\tt\small achilleas.seisa@gmail.com}}
}

\maketitle


\begin{abstract}
This article presents the first real-world autonomous flight of a 5G-connected aerial robot controlled by an edge-offloaded controller, and aims to bridge the gap between controlled and factual setups. The robot operates within an active industrial subterranean mine, while the high-level controller is deployed in a nearby Kubernetes-based edge cluster. Communication between the robot and the edge is enabled via a 5G New Radio (NR) Standalone (SA) network. The chosen controller is a Model Predictive Controller (MPC), which generates control actions to allow the robot to navigate seamlessly through the mining environment. A human operator selects waypoints for the aerial robot, and the MPC generates smooth, collision-free paths for autonomous executions. The proposed 5G edge-based closed-loop system is evaluated in a real industrial setting and demonstrates the potential of edge-controlled robotic systems toward time-critical, safe and efficient future deployments. Related material available in \href{https://drive.google.com/drive/folders/1TbuuloB0kAX3BbS5hdOph9yAdgxdtgZO?usp=sharing}{\texttt{Google Drive}}.
\end{abstract}

\section{Introduction}
\label{sec:intro}
Over the past decades, there has been growing interest in integrating 5G New Radio (NR) Standalone (SA) networks and edge computing technologies for robotic systems applications. However, real-world implementations remain rather limited. Most studies focus on demonstrating 5G-connected and edge-controlled robotic systems in controlled settings, such as simulations or academic laboratory environments. Motivated by the lack of real-world validation and the growing demand for advanced 5G-edge-enabled robotic solutions, this work presents the first deployment of this type for edge-enabled robotics system in active industrial mining environments, with the purpose to validate the feasibility to exploit the current technology to automate some mining operations, leveraging the power of both robotics, as well as of 5G NR SA networks and edge computing.

\begin{figure}[t]
    \centering
    \includegraphics[width=\linewidth]{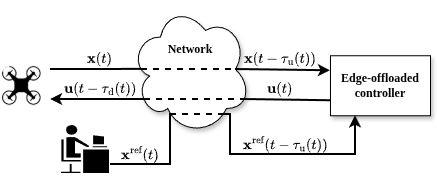}
    \caption{Overview of the proposed 5G-edge framework.}
    \label{fig:5gedge}
\end{figure}

Many studies have investigated this integration of edge computing and sometimes 5G with robotic systems~\cite{groshev2022edge, haidegger2019robotics, 9778241}, as well as the potential of offloading controllers to the edge~\cite{skarin2021explicit, seisa2022edge} to benefit from increased computational resources compared to onboard processors~\cite{robotics7030047} and reduced latency compared to cloud computing~\cite{hu2012cloud}. This concept is extended in the present work to industrial environments, as visualized in Fig.~\ref{fig:5gedge}.

Compared to prior studies conducted in simulation, such as~\cite{zhu2020enabling}, which investigates cloud-based whole-body control of legged robots over a 5G network, and laboratory settings, such as~\cite{10011219}, where Unmanned Aerial Vehicle (UAV) trajectory control is offloaded to the edge via 5G, this work distinguishes itself as the first implementation in an uncontrolled, non-laboratory environment.

For many decades, researchers have aimed to deploy robotic systems in real-world settings~\cite{cui2020offloading}. Robotic systems have been successfully used in several domains such as warehouses~\cite{ikumapayi2024swarm}, manufacturing industries~\cite{xu2021digital, sharma2024edge}, and service applications~\cite{dalgkitsis2023netros}, while adoption in other domains has remained limited. The mining industry, in particular, can significantly benefit from the utilization of autonomous robotic systems, and recent advancements are pushing toward this direction~\cite{obosu2025advances}. From inspection~\cite{szrek2022mobile} and data collection~\cite{du2025industrial} to search and rescue applications~\cite{agha2021nebula}, mining environments are ideal candidates for such solutions. Given that mining operations can be hazardous and difficult to navigate, robots have the potential to greatly enhance both safety and efficiency~\cite{obosu2025advances}. Despite this clear need, robotic systems have not been widely adopted across many industries. In many cases, this is due to the limitations in automation capabilities of these platforms. Advanced algorithms and Artificial Intelligence (AI) aim to revolutionize autonomous systems, but their extensive computational demands often make integration with resource-constrained robotic platforms infeasible.

5G and edge computing have emerged as promising technologies to address these computational limitations in robotic systems~\cite{robotics7030047}. Edge computing provides a platform for offloading computationally heavy applications to infrastructure located at the operational site. When combined with the low-latency communication capabilities of 5G NR SA networks, it becomes possible to meet the real-time requirements of time-critical robotic applications. In this context,~\cite{dalgkitsis2023netros} proposed a 5G-edge-cloud architecture, tested in a laboratory setting, for a service robot that offloads navigation and remote control tasks to edge systems, while video analysis and personalization are handled by cloud servers. Similarly,~\cite{10011219, sankaranarayanan2023paced, damigos2023resilient} explored the use of 5G networks to offload high-level control modules for UAVs. In~\cite{10610698}, a framework is presented for dynamic selection of distinct 5G Quality of Service (QoS) data flows, autonomously managed by a 5G-enabled UAV. In~\cite{cui2020offloading}, researchers proposed offloading computation-intensive autonomous driving services, focusing primarily on application response time rather than communication delays. Furthermore,~\cite{romero2023oros} introduced an edge orchestration approach to minimize mission-critical task completion times and overall energy consumption of 5G-connected robots by jointly optimizing navigation, sensing, and infrastructure resource use. Lastly,~\cite{zahidi2024optimising} proposed an edge-server-over-5G approach for a selective harvesting system.

The motivation for this work arises from the identified gap between recent technological advancements and their real-world adoption, and the disconnection between academic research and industrial application. This study aims to bridge that gap by showcasing a 5G edge-based framework that enables offloaded, time-critical, and safe control operations for robotic platforms in industrial settings, such as subterranean mining sites. The ultimate goal is to contribute to the realization of edge robotics via 5G and to provide practical, real-world solutions. While previous work have proposed edge-based MPC-offloading, this work empowers and demonstrates the real-world realization. 

This work contributes to the domain of edge-controlled robotic systems by presenting, to the best of our knowledge, the first real-world implementation and evaluation of a time-critical, 5G-connected, edge-controlled aerial robotic platform in an uncontrolled, non-laboratory environment. The \textbf{contributions} of this work address key challenges and are centered around: (i) the real-world development of a time-critical closed-loop edge-based framework enabling the offloading of computationally intensive control modules for robotic platforms in industrial settings through Kubernetes (k8s); (ii) the integration of 5G NR SA networks to meet real-time control requirements, and (iii) the validation of the feasibility of such systems in active, operational industrial environments.

The rest of the paper is organized as follows: Section~\ref{sec:problem_formulation} presents the foundational concepts and the recognized challenges of the proposed framework. Section~\ref{sec:system_architecture} describes the system architecture, including the 5G-connected edge-controlled robotic platform, the edge computing setup, and the utilized 5G NR network. Section~\ref{sec:experimental_validation} presents the results obtained from real-world experiments and evaluates the feasibility of edge-controlled platforms in mining environments for safe and efficient deployment. Finally, Section~\ref{sec:conclusions} offers concluding remarks and discusses future developments toward the broader realization of such systems in both industrial and societal applications.

\section{Problem Formulation}
\label{sec:problem_formulation}
While the concept of using edge computing and 5G to offload control modules to edge clusters has been studied, several challenges, such as control stability and communication delays, still hold back real-world implementation.

\subsection{Model Predictive Control}
\label{sec:model_predictive_control}
This work focuses on controlling the trajectory of a UAV based on human waypoint inputs, while simultaneously avoiding collisions. Therefore, we employed Nonlinear Model Predictive Control (NMPC)~\cite{lindqvist2022compra} which is widely adopted by researchers working with UAVs due to its predictive capabilities and high performance. Several studies, such as~\cite{okasha2022design}, have analyzed and compared the performance of UAV controllers, highlighting the superiority of MPC-based control for autonomous missions. This advantage becomes even more critical in industrial and hazardous environments, where human operators or ground robots may be unable to navigate due to complex terrain or safety concerns.

\subsection{Time-Delay Systems}
\label{sec:time-delayed_systems}
The proposed system is a networked control system and is therefore subject to communication delays. Specifically, the state of the robot, denoted as $x(t)$ which incorporates the linear and angular position and velocity information of the robot, is captured at time $t$ and transmitted to the remote controller hosted in the edge k8s cluster. The time required for the state to reach the edge cluster via the uplink (UL) channel is denoted by $t_u$, and thus the state available to the controller is the delayed state $x(t - t_u)$.

The offloaded NMPC controller, running at the edge, computes a control action $u(t)$ which includes the roll, pitch, yaw rates, as well as the thrust, at time $t$, which is then sent to the UAV. The time it takes for this control signal to reach the UAV via the downlink (DL) channel is denoted by $t_d$, resulting in the UAV receiving the delayed command $u(t - t_d)$.

The localization of the robot is estimated on the onboard computer to avoid transmitting high throughput sensor data, such as Light Detection and Ranging (LiDAR) point clouds. Therefore, only telemetry and command actions which carry KB of information are transmitted through the UL and DL channels.

By combining the delays in the UL and DL directions, the round-trip time (RTT) can be defined as $t_{rtt}(t) = t_u(t) + t_d(t) + t_{p}(t)$, where $t_p$ represents the processing time of the MPC controller at the edge.

\subsection{Real-World Implementations}
\label{sec:real_world_implementations}
Cellular networks are known to introduce different delays for UL and DL transmissions, particularly in conventional DL-oriented configurations. This asymmetry arises from both the limited processing capabilities of cellular modems and the use of DL-optimized Time Division Duplex (TDD) patterns, which allocate more resources to DL traffic in order to minimize DL latency relative to UL.

However, control offloading requires robust and low-latency communication in both UL and DL directions to ensure smooth, safe, and stable operation. Therefore, several industrial facilities deploy 5G NR SA infrastructure to enable seamless data offloading to their edge systems. During the experimental phase, both UL and DL delays of the 5G network were measured and incorporated into the MPC model.

Moreover, due to the onboard limitations in executing NMPC modules efficiently, as discussed in~\cite{seisa2024edge}, it is essential to offload the control module to the edge cluster, along with other processes that require significant computational resources. Hence, we can guarantee, that adequate edge computational resources will be allocated for the execution of the controller.

\section{System Architecture}
\label{sec:system_architecture}
The real-world system implementation and architecture consist of three main components: (i) a 5G NR SA network deployed at the industrial mining site, providing reliable radio coverage in the area of operation; (ii) edge computing resources installed on-site and connected to the local 5G NR SA network; and (iii) a 5G-enabled robotic platform, which in this work is a quadcopter drone.

\subsection{5G-Connected Edge-Controlled Robotic Platform}
\label{sec:5g_edge_robotic_platform}
The 5G-enabled robotic platform consists of a quadcopter powered by an onboard \texttt{Intel NUC} companion computer running \texttt{Ubuntu 20.04} and \texttt{ROS 1}, which enables execution of all robotic operations. The drone is equipped with an \texttt{Ouster} LiDAR mounted on the top of the frame, allowing real-time environmental point cloud collection for both positioning and collision avoidance. Cellular modem connectivity and management are handled by the \texttt{ModemManager} package in Linux, while Access Point Name (APN) configuration and connection management are managed via the \texttt{NetworkManager} package.

\begin{figure}[http]
    \centering
    \includegraphics[width=\linewidth]{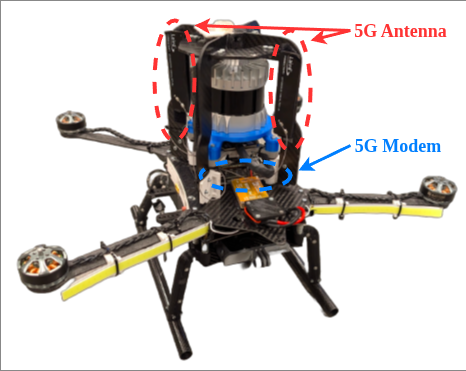}
    \caption{Utilized aerial platform equipped with 5G setup.}
    \label{fig:aerial_platform}
\end{figure}

To enable communication between the aerial robot and the k8s edge cluster, the robot must support 5G connectivity while respecting the system requirements and weight constraints. Given the limited flight time and payload capacity of aerial robots~\cite{dorling2016vehicle}, minimizing onboard weight is critical. To this end, we utilize a custom-made aerial platform~\cite{patel2024towards} equipped with a lightweight 5G communication setup. This setup includes a Sierra Wireless 5G modem (\texttt{EM9293}), an M.2 key B to USB-C adapter, and four antennas positioned 90 degrees apart to provide full spatial coverage. The aerial platform and its 5G setup are illustrated in Fig.~\ref{fig:aerial_platform}.

\subsection{5G Data Transmission Infrastructure}
\label{sec:data_transmission}
The 5G network installation at the industrial site is composed by \texttt{Ericsson} base stations adapted for the harsh conditions of the mining environment. The outdoor coverage is provided by a standard multi-sector macro base station setup operating in the 3.5 GHz spectrum (band N78). For indoor coverage within the tunnel-like environment, base stations are customized with a two-sector configuration instead of the typical three-sector setup. This is combined with highly directive antennas to ensure better longitudinal coverage along tunnels and effective radio overlap between adjacent sites. In contrast, wider open areas utilize more omnidirectional radio equipment to provide broader coverage.

All radio equipment is connected via optical fiber to underground Remote Radio Units (RRUs), which in turn are linked to a centralized Base Band Unit (BBU), also located underground. The 5G network core, along with the edge computing infrastructure, is instead located above ground in a dedicated enclosure.

The network installation includes multiple 5G slices. One slice is dedicated to the experimental testbed and optimized for low-latency applications, while another is configured for higher-latency use cases, such as providing basic connectivity for workers’ mobile devices and other non-critical mining equipment.

\subsection{Edge-Offloaded Control}
\label{sec:edge-offloaded_control}
The resources area accessible through a containerization and Virtual Machine (VM) system executed over a virtualization Operating System (OS) running on the bare metal machine, which allows to properly allocate and exploit the hardware resources both to the proper network slice, and to the demanding application requirements.

The edge resources are deployed as a k8s cluster composed of four active nodes based on an Intel multi-core Central Processing Unit (CPU), aided by 590 GiB of Random Access Memory (RAM) and \texttt{NVIDIA} Graphics Processing Unit (GPU) capabilities (\texttt{NVIDIA GTX 2080 Ti}) for more advanced Machine Learning (ML) and Artificial Intelligent (AI) processing related tasks. The cluster is managed through Rancher Kubernetes Engine (RKE) and runs k8s version 1.28.15. The nodes in the cluster are a mix of control plane and worker nodes. This setup, along with dynamic resource allocation as described in~\cite{seisa2024cloud}, can guarantee faster and optimized MPC execution~\cite{seisa2024edge}.

Access to the resources is managed through containerized environments and VM running atop a virtualization layer on bare-metal hardware. This setup enables fine-grained resource allocation aligned with both the demands of specific applications and the constraints of network slicing, supporting real-time and compute-intensive workloads across the edge infrastructure.

\begin{figure}[http]
    \centering
    \includegraphics[width=0.8\linewidth]{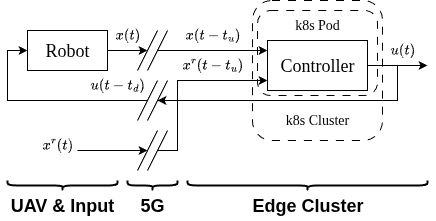}
    \caption{Block diagram of the edge-offloaded closed-loop control system.}
    \label{fig:arch}
\end{figure}

The overall architecture is illustrated in Fig.~\ref{fig:arch}. The aerial robot transmits sensor data to the edge k8s cluster via the 5G NR SA connection. Within the cluster, the NMPC is deployed as a containerized application running in a dedicated k8s pod. The pod is configured with predefined NodePort services to facilitate seamless bidirectional communication over User Datagram Protocol (UDP), enabling efficient transmission of telemetry and sensor data from the robot to the edge, and control commands from the NMPC back to the robot.

The NMPC container is deployed using Rancher and is managed under the namespace drone-controller, with specific port mappings (e.g., 30200–30203) to expose the relevant endpoints externally. The communication stack is fully containerized, ensuring that the control loop can operate in near real-time while taking full advantage of the low latency and high throughput provided by the 5G network and edge computing infrastructure.

\section{Experimental Validation}
\label{sec:experimental_validation}
\begin{figure}[httb]
    \centering
    \includegraphics[width=\linewidth]{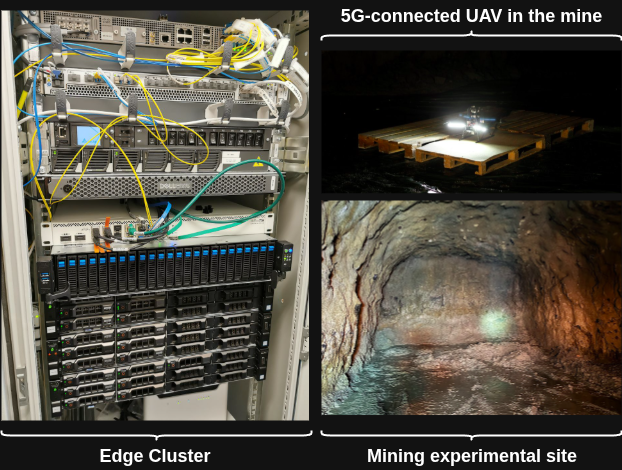}
    \caption{Experimental setup including the edge cluster, the mining site and the 5G-connected aerial robot.}
    \label{fig:setup}
\end{figure}

The concept of edge-offloaded control has been validated in simulation, laboratory, and real-world scenarios, with video and slide demonstrations available in the accompanying \href{https://drive.google.com/drive/folders/1qRP4dFaWzTpC6ebHVofonmd-8L_pBDAa?usp=sharing}{\texttt{Google Drive}}. After thorough evaluation in controlled environments, experiments were conducted at an active mining site in northern Sweden. The aerial robot, equipped with all necessary modules, was deployed in the mine, which featured infrastructure supporting both 5G networks and on-premises edge computing. The edge cluster consisted of a k8s environment managed using Rancher, which is an open-source multi-cluster orchestration platform. The experimental setup is depicted in Fig.~\ref{fig:setup}.

The experimental scenario involved a human operator who connected to the edge cluster to select waypoints for the robot to navigate. The edge-offloaded controller received this input along with sensor data transmitted from the robot and generated control actions to enable the robot to seamlessly follow the designated waypoints.

\begin{figure}[httb]
    \centering
    \includegraphics[width=\linewidth]{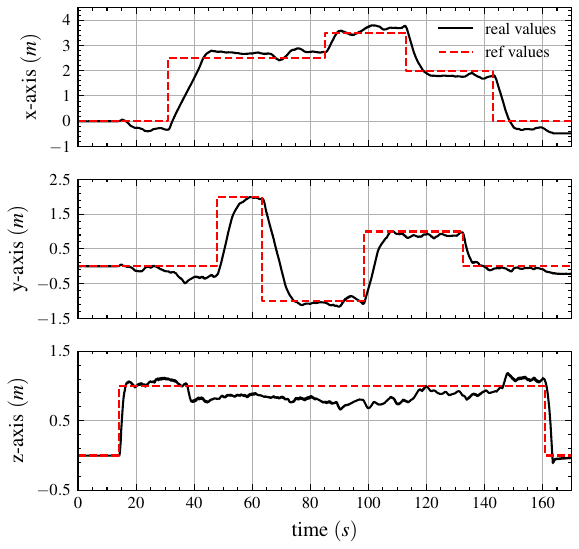}
    \caption{Robot's real trajectory and reference waypoints in the $x$, $y$, and $z$ planes.}
    \label{fig:trajectory}
    \vspace{-0.2cm}
\end{figure}

System performance is evaluated in Fig.~\ref{fig:trajectory}, which illustrates the human-selected waypoints and the actual trajectory of the aerial robot. The figure demonstrates the robot’s ability to accurately follow the waypoints while its controller is offloaded to an edge-based k8s cluster.

In Table~\ref{tab:network}, we present the 5G network radio specifications and measurements collected during the experimental deployment. The table includes details such as the operating band, bandwidth (DL and UL), and key radio parameters including Reference Signal Received Power (RSRP), Signal-to-Interference-plus-Noise Ratio (SINR), TX Power (transmit power), and per-antenna Received Signal Strength Indication (RSSI) values. These measurements provide insights into the link quality and signal propagation conditions within the subterranean industrial environment.

\begin{table}[ht!]
    \centering
    \caption{5G Network Specifications \& Measurements}
    \label{tab:network}
    \resizebox{\columnwidth}{!}{%
    \begin{tabular}{ccc}
    \toprule
    \textbf{Parameter} & \textbf{Minimum Value} & \textbf{Maximum Value}\\
    \midrule
    Band & \multicolumn{2}{c}{n78}\\
    BW & \multicolumn{2}{c}{100 MHz}\\
    RSRP & -100 dBm & -83 dBm\\
    SINR & 12.5 dB & 27.0 dB\\
    EU TX Power & 13.0 dBm & 23.0 dBm\\
    RSSI Ant 0 & -72.3 dBm & -43.3 dBm\\
    RSSI Ant 1 & -71.9 dBm & -45.3 dBm\\
    RSSI Ant 2 & -73.8 dBm & -55.9 dBm\\
    RSSI Ant 3 & -69.1 dBm & -49.1 dBm\\
    \bottomrule
    \end{tabular}%
    }
\end{table}

To assess the end-to-end network performance, we further evaluated the communication delay and jitter introduced by the 5G NR SA network. For this purpose, we used \texttt{Qosium}, a real-time passive network monitoring and performance analysis tool. The master measurement probe has been set on the edge server, while the secondary probe on the companion computer board installed on the drone. The measurements captured include the maximum delay, drift-corrected delay, and network jitter, which are critical metrics for validating the feasibility of time-sensitive control applications. These measurements provide quantitative insight into the communication reliability and timing stability of the 5G link under real operating conditions.

\begin{figure}[httb]
    \centering
    \includegraphics[width=\linewidth]{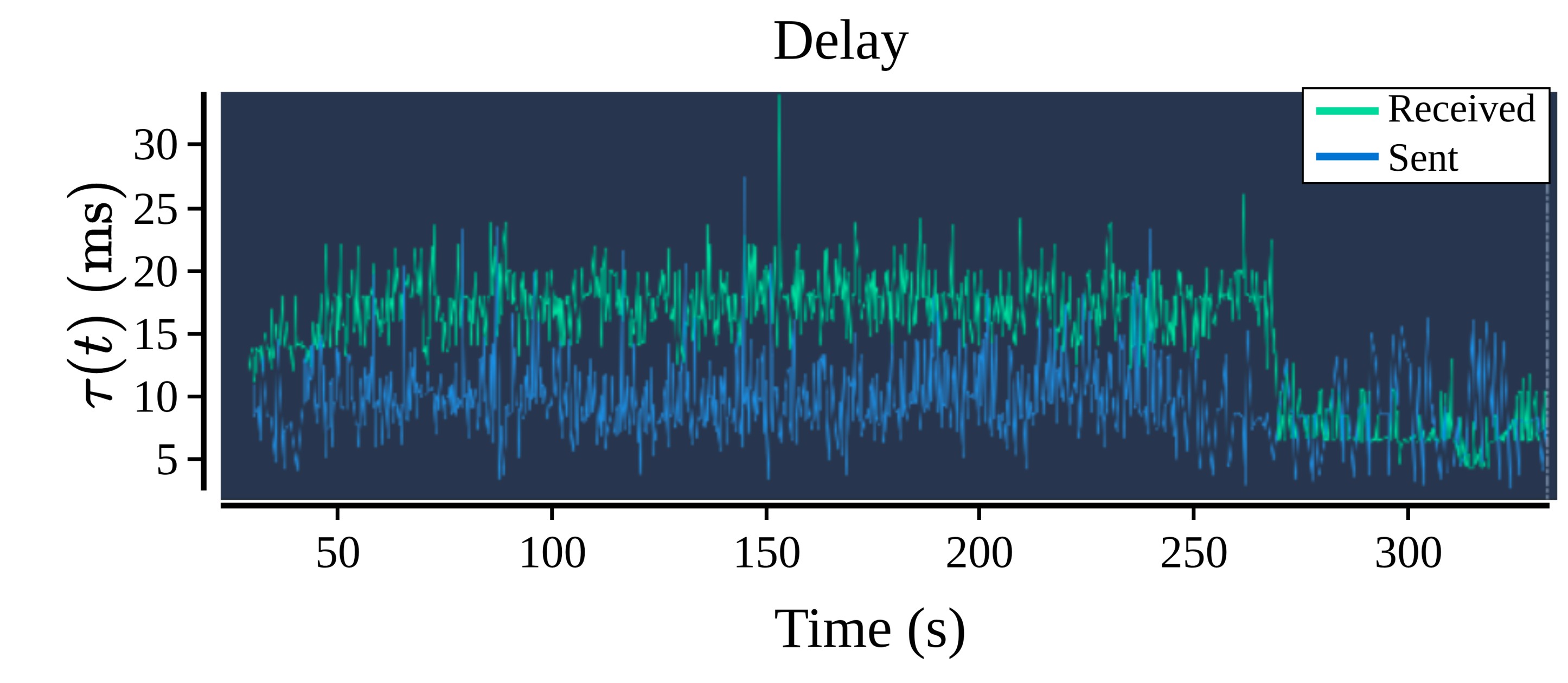}
    \caption{Maximum network delay for DL (blue) and UL (green) channels.}
    \label{fig:delay}
\end{figure}

Fig.~\ref{fig:delay} illustrates the measured maximum network delay between the aerial robot and the edge cluster. This value represents the highest observed sample delay within each averaging interval. The UL delay ranges between 5 ms and 34 ms, while the DL delay varies from 3 ms to 27 ms.

\begin{figure}[httb]
    \centering
    \includegraphics[width=\linewidth]{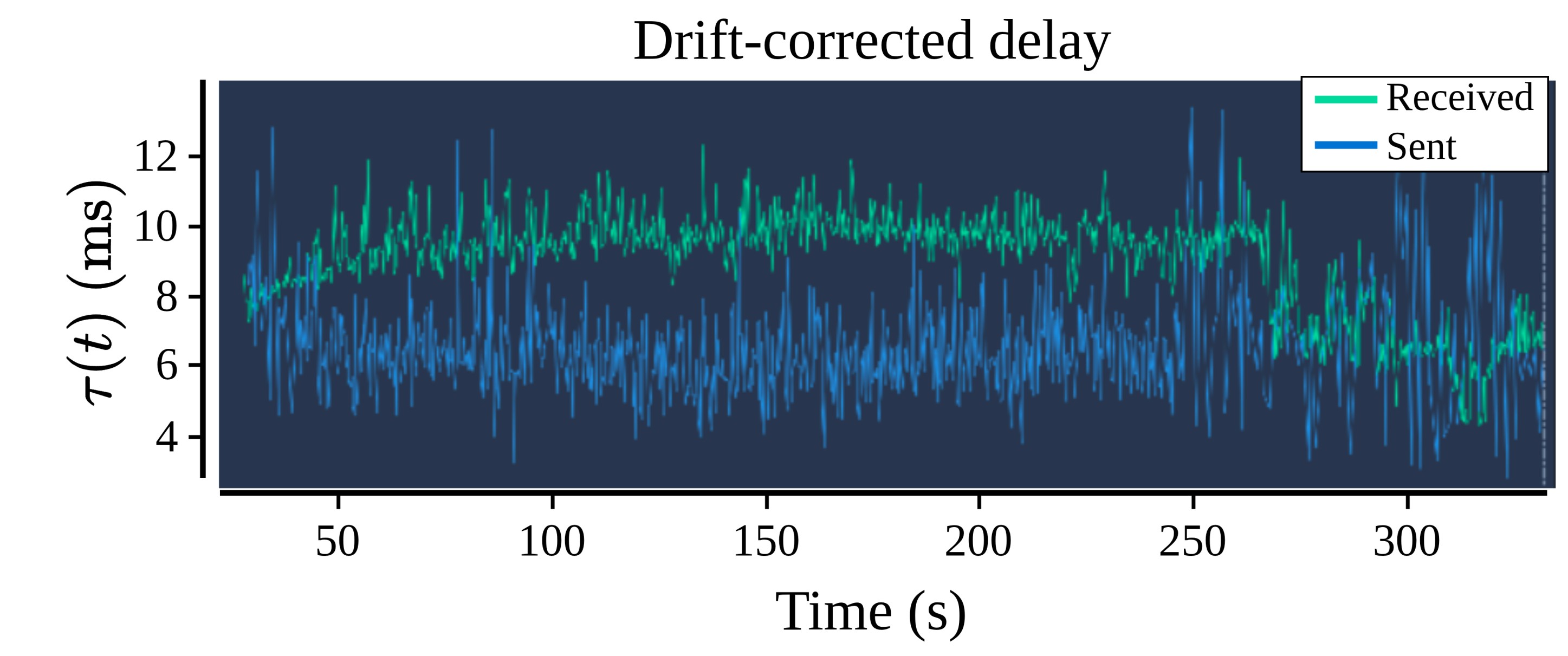}
    \caption{Drift-corrected delay for DL (blue) and UL (green) channels.}
    \label{fig:delay_corrected}
\end{figure}

Fig.~\ref{fig:delay_corrected} shows the drift-corrected delay, which accounts for potential clock drift between the measurement endpoints and provides an estimate of the actual communication delay. The corrected UL delay ranges from 4 to 13 ms, with an average of 9.5 ms, and the DL delay ranges from 2 to 15 ms, with an average of 6.5 ms.

These delay values confirm that the closed-loop system operates within a stable regime. According to~\cite{seisa2024edge}, system stability is maintained as long as the RTT satisfies the condition:
\begin{align}
        t_{rtt}(t)<\tau_{max}, \nonumber
\end{align}
where $\tau_{\text{max}}$ denotes the maximum allowable delay for stable operation under the offloaded NMPC control architecture.

\begin{figure}[httb]
    \centering
    \includegraphics[width=\linewidth]{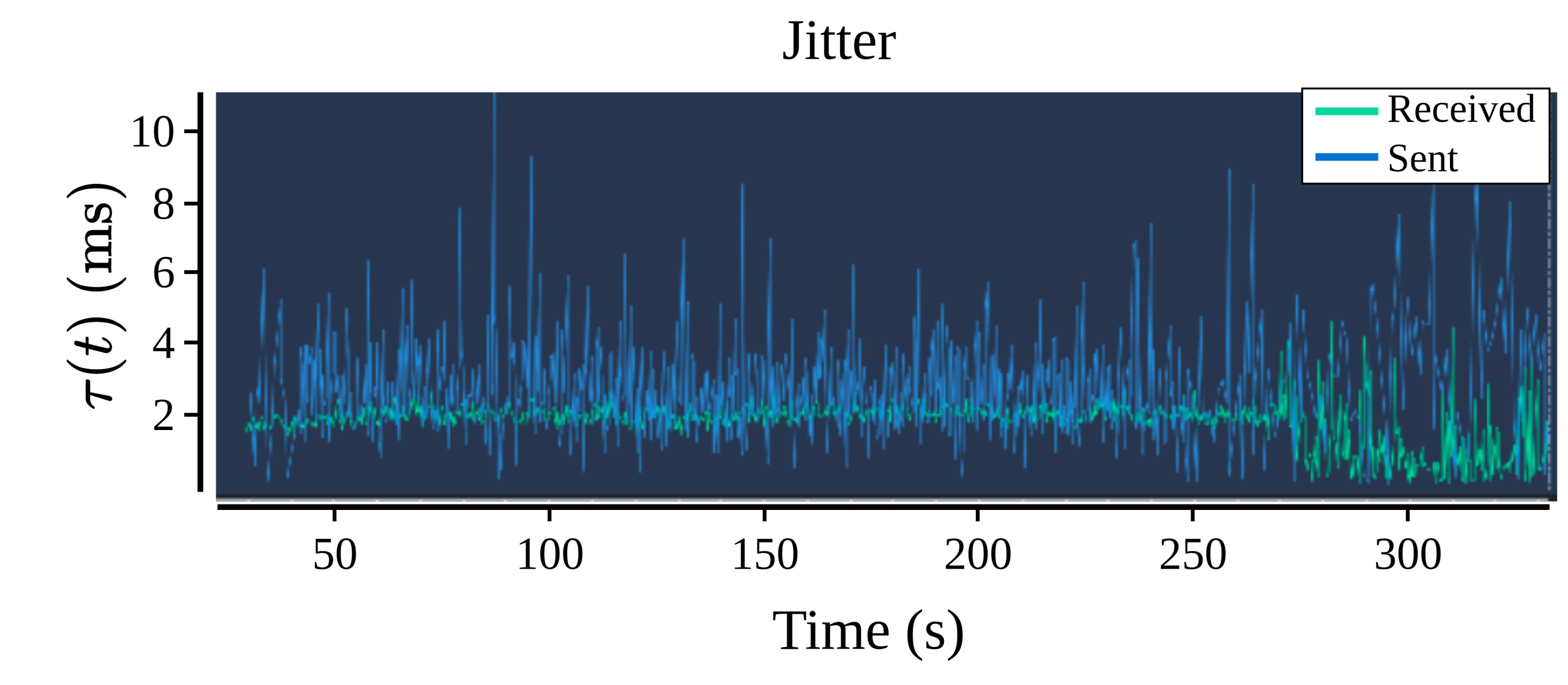}
    \caption{Network jitter for DL (blue) and UL (green) channels.}
    \label{fig:jitter_act}
\end{figure}

Fig.~\ref{fig:jitter_act} shows the network jitter, i.e., the pure delay variation for DL and UL channels. The jitter values range from 0 to 5 ms and from 0 to 11 ms for the UL and DL directions, respectively. Finally, the measured average UL traffic was 2.1 Mbps, while the DL was 0.1 Mbps.

\section{Conclusions and Future Work}
\label{sec:conclusions}
The experimental results of this work demonstrate that the current capabilities of advanced 5G cellular networks can be effectively exploited for industrial automation in harsh environments. They also show how real-time data processing at the edge can reduce the need for onboard computational resources, making robotic platforms more affordable, lightweight, and easily upgradable. By shifting the processing load to the edge, system maintenance becomes more seamless and cost-effective.

Future developments of the proposed architecture may involve more advanced data processing offloading, enabled by UL-oriented 5G networks and more powerful edge infrastructure. Furthermore, with respect to mission planning for autonomous robots using offloaded controllers, new paradigms can be considered—such as incorporating network link quality parameters into the planning process. These parameters could inform decisions about the amount of data to offload and the operational distance limits of robotic missions, ultimately contributing to safer and more reliable deployments in challenging environments.

\bibliographystyle{./IEEEtranBST/IEEEtran}
\bibliography{./IEEEtranBST/IEEEabrv, references}

\end{document}